\documentclass{article}

\usepackage[preprint]{neurips_2024}

\usepackage{wrapfig}
\usepackage{subfig}
\usepackage{algorithm}
\usepackage{algpseudocode}
\usepackage[dvipsnames]{xcolor}
\usepackage{mathtools}
\usepackage{lipsum}
\usepackage{color, colortbl}
\usepackage{amsmath}
\usepackage{graphics}
\usepackage{listings}
\definecolor{codegreen}{rgb}{0,0.6,0}
\definecolor{codegray}{rgb}{0.5,0.5,0.5}
\definecolor{codepurple}{rgb}{0.58,0,0.82}
\definecolor{backcolour}{rgb}{0.95,0.95,0.92}

\usepackage[utf8]{inputenc} 
\usepackage[T1]{fontenc}    
\usepackage[backref=page]{hyperref}       
\usepackage{url}            
\usepackage{booktabs}       
\usepackage{amsfonts}       
\usepackage{nicefrac}       
\usepackage{microtype}      
\usepackage{xcolor}         
\title{\textsc{LoGAH}: Predicting 774-Million-Parameter Transformers using Graph HyperNetworks with \nicefrac{1}{100} Parameters}

\author{%
  Xinyu Zhou \\EPFL  \And
  Boris Knyazev \\ Samsung - SAIT AI Lab \And
  Alexia Jolicoeur-Martineau\\ Samsung - SAIT AI Lab \And
  Jie Fu\thanks{Corresponding Author.} \\HKUST \AND
}

\begin{document}

\maketitle


\newcommand{\theHalgorithm}{\arabic{algorithm}}

\newcommand{\boris}[1]{\textcolor{blue}{Boris: #1}}








\begin{abstract}
A good initialization of deep learning models is essential since it can help them converge better and faster. 
However, pretraining large models is unaffordable for many researchers, which makes a desired prediction for initial parameters more necessary nowadays. Graph HyperNetworks (GHNs), one approach to predicting model parameters, have recently shown strong performance in initializing large vision models. 
Unfortunately, predicting parameters of very wide networks relies on copying small chunks of parameters multiple times and requires an extremely large number of parameters to support full prediction, which greatly hinders its adoption in practice.
To address this limitation, we propose \textsc{LoGAH} (Low-rank GrAph Hypernetworks), a GHN with a low-rank parameter decoder that expands to significantly wider networks without requiring as excessive increase of parameters as in previous attempts.
\textsc{LoGAH} allows us to predict the parameters of 774-million large neural networks in a memory-efficient manner. We show that vision and language models (i.e., ViT and GPT-2) initialized with \textsc{LoGAH} achieve better performance than those initialized randomly or using existing hypernetworks. 
Furthermore, we show promising transfer learning results w.r.t. training \textsc{LoGAH} on small datasets and using the predicted parameters to initialize for larger tasks. We provide the codes in \url{https://github.com/Blackzxy/LoGAH}.

\end{abstract}

\section{Introduction}
\label{intro}
In vision and language domains,
pretraining a large model from scratch precedes solving downstream tasks~\citep{he2021masked, devlin-etal-2019-bert}.
Recent models have been increasing in size dramatically, chasing state-of-the-art performance: from around 100M to $\geq$65B parameters for Generative Pretrained Transformers (GPTs)  \citep{radford2018improving,touvron2023llama,llama3modelcard} and from around 100M to $\geq$22B for Vision Transformers (ViTs) \citep{dosovitskiy2021image,dehghani2023scaling}. 
Training such large models requires large computing resources. In addition, many retraining iterations are often required before the model is successfully trained, which is exacerbated in larger models since they are often more unstable to train and require more hardware and software tuning in addition to hyperparameter and architecture tuning, data curation, etc. 
Thus, pretraining large models has become very expensive even for big companies~\citep{thompson2022computational, zhai2022scaling}. 
\begin{figure}
  \centering
  \subfloat[Comparison of parameter counts between GHN-3 and \textsc{LoGAH} for supportable maximal widths in the predicted parameters (without their copying).]{\includegraphics[width=0.47\textwidth]{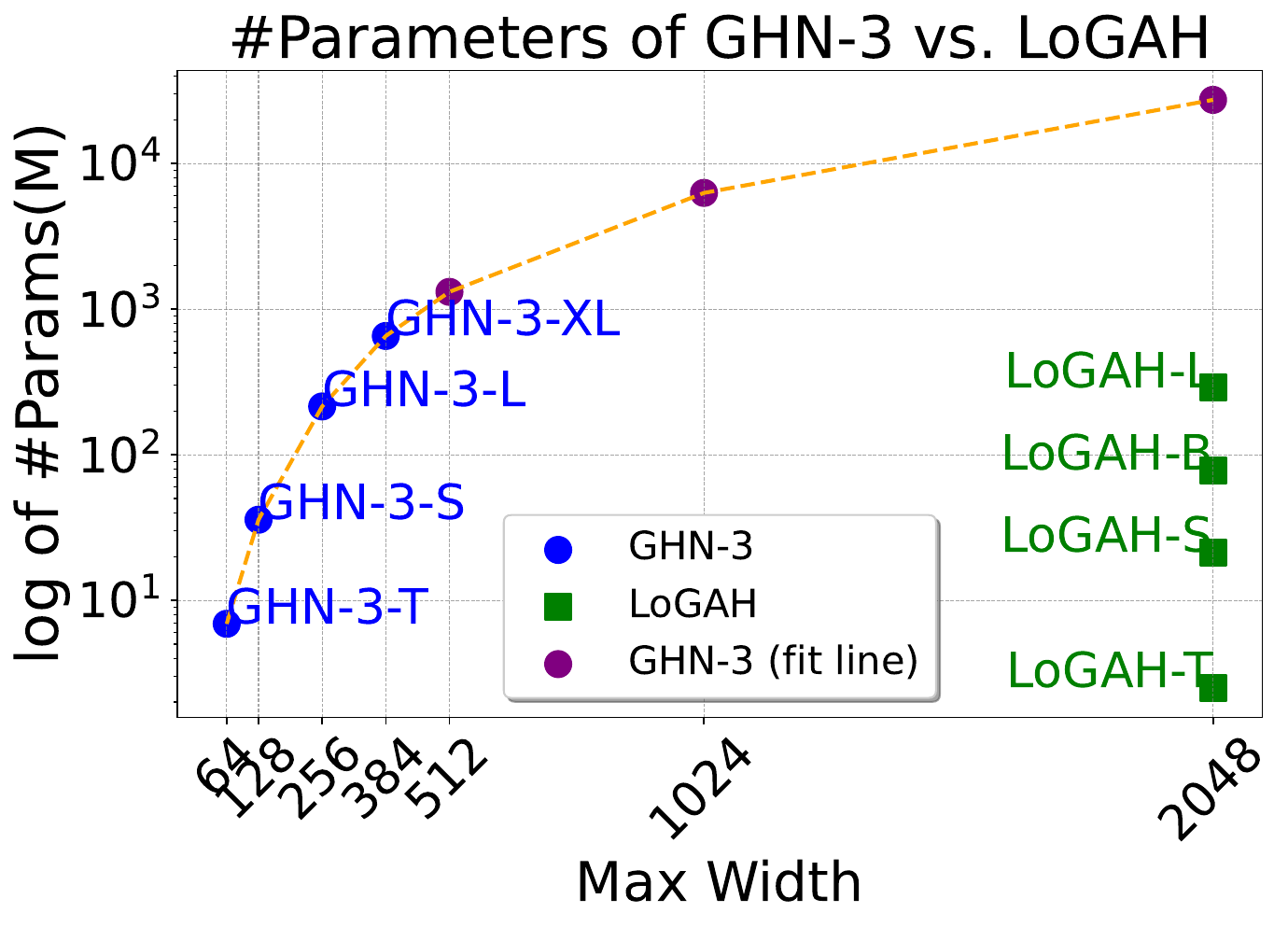}\label{param_width}}
  \hfill
  \subfloat[Comparison of parameter counts  between GHN-3 and \textsc{LoGAH} for supportable network sizes (without copying predicted parameters).]{\includegraphics[width=0.47\textwidth]{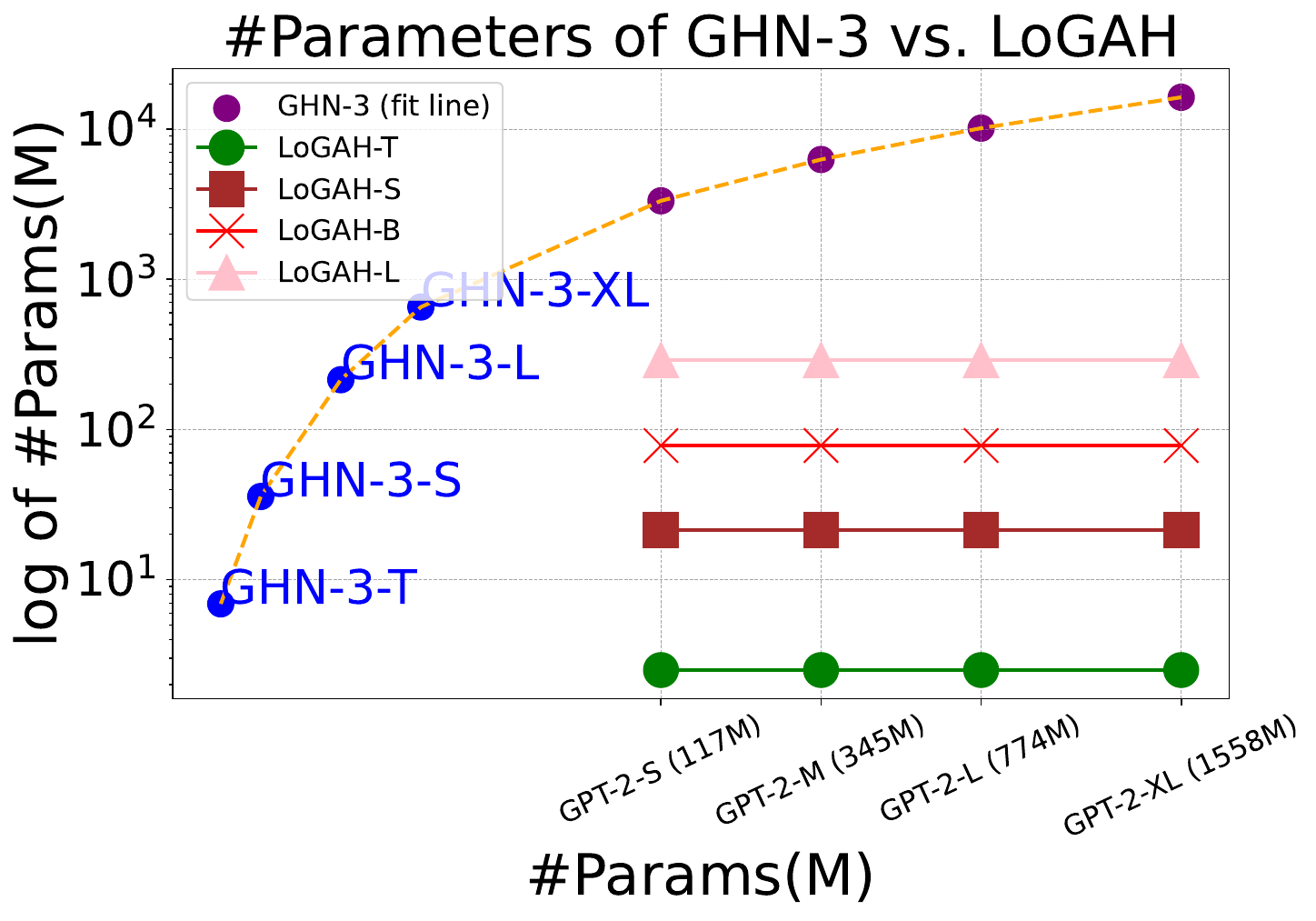}\label{param-comp}}
  \caption{Comparison of parameter counts between GHN-3 and \textsc{LoGAH}. GHN-3 requires a larger hidden size to support wider networks, which increases the size of GHN-3 exponentially in Figure \ref{param_width}. \textsc{LoGAH} can support much wider networks (up to 2048-dimension), and larger networks (GPT-2-Large in 1280-dimension with 774M parameters) even using \textsc{LoGAH-Tiny}.}
  \label{params-comp-2ghn}
\end{figure}

Currently, training vision and language tasks are generally done using similar network architectures and datasets;
architectures are generally based on either Transformers~\citep{vaswani2023attention} (for vision or language) or Convolutional Neural Networks (CNNs) \citep{fukushima1983neocognitron} (for vision), while datasets are similar to either ImageNet (for vision)~\citep{russakovsky2015imagenet} or The Pile (for language)~\citep{gao2020pile}. Leveraging this prior knowledge of the architecture and dataset may reduce the pretraining cost. One approach to do so is Graph HyperNetworks (GHNs)\citep{zhang2018graph,knyazev2021parameter, knyazev2023scale}; this approach allows one to predict initial parameters of these neural networks that perform well and converge faster. We describe the GHN approach below.

Using a set of neural network architectures $\{f^G\}$ as training data, GHN $H_{\mathcal{D}}$, parameterized by $\theta$, is trained to predict the parameters of these neural networks ($\textbf{w}_{\text{pred}} = H_{\mathcal{D}}(f^G, \theta)$) to minimize the loss function on the dataset $\mathcal{D}$.
%
The predicted $\textbf{w}_{\text{pred}}$ can serve as a stronger initialization compared to random-based initialization methods, thus greatly reducing the pretraining cost. 

However, to predict parameters for very wide networks (often with a large number of parameters), previous GHNs \citep{knyazev2021parameter, knyazev2023scale} had to copy small chunks of parameters multiple times instead of fully predicting them due to the sheer amount of parameters required to predict all parameters, thus significantly limiting the performance of the resulting networks. Furthermore, to unlock the capability of predicting parameters of a larger size, GHNs need larger hidden sizes $d$, leading to an exponential increase in the number of parameters growing as $\mathcal{O}(d^3)$ (Figure \ref{param_width}).

To overcome this limitation, we propose \textsc{LoGAH}, a GHN with a low-rank parameter decoder. This novel approach not only supports significantly wider networks but also does so without requiring an excessive number of parameters growing as $\mathcal{O}(d^2)$ instead of $\mathcal{O}(d^3)$ (Figure~\ref{param-comp}). For instance, our smallest \textsc{LoGAH-Tiny} has only 2.5M parameters, yet it can predict parameters with up to 2048 channels, including GPT-2-Large with 774M parameters and potentially even larger networks.

In this work, we make the following contributions: 
\begin{itemize}
    \item We propose \textsc{LoGAH}, with an improved low-rank decoder, that is more scalable and can predict parameters of large networks without copying while having fewer trainable parameters and a lower training cost (Section \ref{logah}).
    \item We create a new dataset of small ViT and GPT-2 architectures, allowing GHNs to be trained on Transformers for both vision and language domains (Section \ref{vit-gpt-dataset}). \textsc{LoGAH} shows excellent generalized capability on larger models.
    \item We outperform GHN-3 as an initialization approach in multiple vision and language tasks by predicting more diverse and performant parameters (Section \ref{experiments}).
\end{itemize}

\section{Preliminaries}
\subsection{Graph HyperNetworks}
Graph HyperNetworks (GHNs) \citep{zhang2020graph, knyazev2021parameter} are widely used for neural networks' parameter prediction. The input fed to GHN $H_{\mathcal{D}}(\theta)$ is a computational graph $f^G$ of a neural network $f$; GHN predicts its parameters $\textbf{w}_{\text{pred}}=H_{\mathcal{D}}(f^G;\theta)$, where $\mathcal{D}$ is the training dataset. In our paper, $f$ can be a ViT model \citep{dosovitskiy2021image} (\textit{resp.} GPT-2 \citep{radford2019language}), and $\mathcal{D}$ can be the image classification task (\textit{resp.} causal language modeling task). 

In \citet{knyazev2021parameter} work, GHN $H_{\mathcal{D}}$ is trained by SGD over $M$ training architectures $\{f_{a}^G\}_{a=1}^M$ and $N$ training data samples $\{\textbf{x}_j, y_j\}_{j=1}^N$ on the following optimization problem:
\begin{align}
    \arg \min_{\theta} \frac{1}{NM}\sum_{j=1}^N\sum_{a=1}^M\mathcal{L}(f_a(\textbf{x}_j;H_{\mathcal{D}}(f_{a}^G;\theta)), y_j).
\end{align}
A meta-batch of $m$ training architectures is sampled in the training stage where $H_{\mathcal{D}}$ predicts parameters. 
Meanwhile, a mini-batch of $n$ training samples $\textbf{x}$ is sampled and fed into the parameter-predicted $m$ architectures to get $m\times n$ predictions. The cross-entropy loss $\mathcal{L}$ is computed for classification and language modeling tasks (next-token prediction). Afterward, the loss is back-propagated to update the parameters $\theta$ of $H_{\mathcal{D}}$ by gradient descent. In our work, we created \textsc{ViTs-1K} and \textsc{GPTs-1K} datasets of small training architectures for predicting parameters for larger ViT and GPT-2 models, respectively. We describe the details in Section \ref{vit-gpt-dataset}.

The computational graph $f^G=(V,E)$ for input is a Directed Acyclic Graph (DAG), where $V$ denotes the operations (e.g., pooling, self-attention, etc.), and $E$ corresponds to the forward pass flow of inputs through $f$. The $d$-dimensional node features $\textbf{H}^{(1)}\in \mathbb{R}^{|V|\times d}$ are obtained by an embedding layer ($i$-th node: $\textbf{h}_i^{(1)}=\text{Embed}(\textbf{h}_i^{(0)})$, where $\textbf{h}_i^{(0)}$ is a one-hot vector representing for an operation) and fed as the input for GHN. After $L$ Graphormer layers \citep{ying2021transformers}, the node features $\textbf{H}^{(L)}\in \mathbb{R}^{|V|\times d}$ are fed to the decoder described below.

\subsection{GHN Decoder}


\citet{knyazev2021parameter,knyazev2023scale} have the decoder based on a simple MLP predicting a tensor of shape $d\times d\times 16 \times 16$, where $d$ is relatively small ($d=384$ even in the largest GHN-3).  The decoder takes the output node features of the last Graphormer layer to predict parameters $\textbf{w}_{\text{pred}}$. This tensor is copied when the target weight has a larger $d$ or sliced when the target is smaller. The parameter count of the decoder in \citep{knyazev2021parameter,knyazev2023scale}\footnote{Please refer to Appendix \ref{apdx: ghn3-decoder} and \url{https://github.com/SamsungSAILMontreal/ghn3/blob/main/ghn3/nn.py} for more details} is:
\begin{align}
    \#\text{Param}_{\text{GHN-decoder}} = 4d^2\times 16\times 16+32d^2+8d^3+d\times \text{num\_class} \in \mathcal{O}(d^3).
\label{num_param_decoder1}
\end{align}

\section{Scalable Graph HyperNetworks: \textsc{LoGAH}}
\label{logah}

\textsc{LoGAH} model improves on the following aspects: (1) designing a novel low-rank decoder not only with fewer amounts of parameters, but also avoiding inefficient parameter repetitions on prediction, (2) supporting larger models (often wider) prediction without involving extremely larger amounts of parameters as in previous works, e.g. \textsc{LoGAH-Tiny} with only 2.5M parameters can support GPT-2-Large, while existing methods~\citep{knyazev2023scale} would require at least $\sim 10^5$M parameters.

\subsection{Low-Rank Decoder}
In \citep{knyazev2023scale}, the final output dimensionality of the decoder is $d\times d\times 16 \times 16$, where $d$ can be $64$ or $128$. 
In most cases, $16\times 16$ can be a waste since convolutional parameters are generally in $3\times 3$ or $7\times 7$. However, the bigger problem is that for large networks, the tensor needs to be repeated to fill all channels because $d$ is small.

Considering a convolutional weight $W$ with size: $(C_{out} \times C_{in}\times h\times w)$, we can reshape it into a matrix $W$ of $(C_{out}\cdot h)\times (C_{in}\cdot w)$ where $h, w$ are much smaller than $C_{out}$ and $C_{in}$. Inspired by~\citep{hu2021lora}, we can now introduce the low-rank decomposition:
\begin{align}
    W = AB\in\mathbb{R}^{(C_{out}\cdot h)\times (C_{in}\cdot w)},
    \label{low-rank-decomp}
\end{align}
where $A \in\mathbb{R}^{(C_{out}\cdot h)\times r}, B\in \mathbb{R}^{r\times (C_{in}\cdot w)}$, $r$ denotes the low-rank. In this way, we reduce the amounts of parameters from $C_{out}\cdot C_{in}\cdot h\cdot w$ to $r\cdot ((C_{out}\cdot h) + (C_{in}\cdot w))$. 

Therefore, the whole process is as follows: after the first MLPs (multilayer perceptron) the input $\textbf{H}^{(L)}\in \mathbb{R}^{|V|\times d}$ is transformed into $\Tilde{W}\in \mathbb{R}^{|V|\times 2K\times r}$:
\begin{align}
    \Tilde{W}=\text{MLP}(\textbf{H}^{(L)})\in\mathbb{R}^{|V|\times 2K\times r},
\end{align}
where $K:=\max(C_{out}\cdot h, C_{in}\cdot w)$ is called \textbf{max mask}, so that we can avoid repetition operations in \textsc{GHN-3}. Then we split $\Tilde{W}$ into two matrices $A, B^T\in\mathbb{R}^{|V|\times K\times r}$ and only take the needed bits to construct $W = AB$ in Eqn. (\ref{low-rank-decomp}). The architecture of the MLPs is shown in Appendix \ref{mlps_ghn_lora}, which involves the low-rank transformation inside. In this way, the number of parameters in the decoder of \textsc{LoGAH} is:
\begin{align}
    \#\text{Param}_{\text{LoGAH-decoder}} = 4d^2+32d^2+8d\times 2r^2+r\times K.
    \label{num_params_decoder2}
\end{align}

Theoretically, we can fix $r$ as a much smaller constant hyperparameter than $d$, then Eqn. (\ref{num_params_decoder2}) would be in $\mathcal{O}(d^2)$, less than the complexity of original GHN's decoder $\mathcal{O}(d^3)$. In practice, considering a small rank $r$ would hinder the model's performance, so we set it to $r\approx \frac{d}{2}$ as an increase of $d$. Under this setting, we compare the amounts of two decoder's parameters in detail as follows.

\textbf{\#Parameters Comparison.} Without loss of generality, we assume $K=C_{out}\cdot h$, and in our following settings for low-rank $r$ (details in Table \ref{ghn_lora_versions})\footnote{Although in \textsc{LoGAH-Large} setting: $d=r=256$, Eqn. (\ref{1st_last_dif}) will obtain $16d\cdot(64d-2048)>0$ since $d$ is very large. We also tried $d=384, r=256$, however, the training is unstable.}:$r \approx \frac{d}{2}$. Then Eqn. (\ref{num_param_decoder1}) -  Eqn. (\ref{num_params_decoder2}) we obtain:
\begin{align}
    \Delta\textbf{P} = \textcolor{ForestGreen}{4d^2\times(16^2-1)}&+8d\times(d^2-2r^2)+d\times \text{num\_class} \textcolor{OrangeRed}{-r\times C_{out}\cdot h}.
    \label{param_dif}
\end{align}

Since $r\approx d/2$, $16^2-1\approx 16^2$, and in our experiments we set $K=\max(C_{out}\cdot h, C_{in}\cdot w)=2048\cdot 16$, we can just compare the \colorbox{YellowGreen}{first term} and \colorbox{Salmon}{last term} in Eqn . (\ref{param_dif}):
\begin{align}
    \Delta_1 &= 4d^2\times(16^2-1)-r\times C_{out}\cdot h\\
            &\approx 4d^2\times 16^2-d\times 1024\cdot 16\\
            &=16d\cdot(64d-1024). \label{1st_last_dif}
\end{align}
Therefore, $\Delta_1 >0$ since in our settings $d=64, 128, 256$, etc, which means that \textsc{LoGAH}'s decoder requires fewer parameters ($\Delta\textbf{P}>0$), even if we let $r$ increase with $d$.

\subsection{Predicting parameters in larger shapes with fewer parameters}
Thanks to the low-rank mechanism, \textsc{LoGAH} can support predicting the parameter tensors with a larger shape but with fewer parameters. The parameters comparison between different versions of \textsc{GHN-3} and \textsc{LoGAH} is shown in Figure \ref{params-comp-2ghn}. Since \textsc{GHN-3} can only support the predicted parameters as the same width as the hidden size $d$, we fit the curve of GHN-3 and obtain the potential number of parameters needed to fully predict parameters with larger shapes. Compared to GHN-3, our \textsc{LoGAH} can support wider tensor shapes with much fewer parameters, which can support larger models (often wider) in practice (referring to Table \ref{vit-version} and Table \ref{gpt2-version}).




\section{\textsc{ViTs-1K} and \textsc{GPTs-1K} Datasets}
\label{vit-gpt-dataset}
For sampling training architectures in previous GHN-related works, \cite{knyazev2021parameter} built DeepNets-1M, a dataset of 1 million diverse computational graphs. However, for generating Transformer models such as ViT and GPT-2, DeepNets-1M is not optimal. Therefore we introduce \textsc{ViTs-1K} and \textsc{GPTs-1K}: these new datasets contains 1K different ViT-style and GPT-2-style computational graphs respectively, particularly for training GHNs to predict ViT and GPT-2's parameters. 

\textbf{\textsc{ViTs-1K}.}
We produce diverse ViT models by varying the number of layers $L$, heads $H$ and hidden size $D$. Since ViT models have different scale versions (as illustrated in Table \ref{vit-version} of Appendix \ref{apdx:vit-gpt-details}), we also need to ensure that our training architectures will be diverse enough and uniformly distributed in terms of parameter count. Therefore, when generating these architectures, for deeper networks (with more layers) we control them to be narrower (with a smaller hidden size) and vice versa. Figure \ref{vits1k-params-dis} shows the distribution of the amounts of parameters in \textsc{ViTs-1K}, which is almost uniformly distributed and the maximum parameters of these architectures are restricted to 10M (only around of half of ViT-Small's parameters). The details of \textsc{ViTs-1K} dataset's generation can be found in Appendix \ref{appendix_vits1k}.

\textbf{\textsc{GPTs-1K}.}
We follow the same above idea to get different GPT-2 models, by varying the number of layers $L$, heads $H$ and hidden size $D$, to build \textsc{GPTs-1K}. The parameter count distribution is shown in Figure \ref{gpt2s1k-params-dis}, and the maximum parameter count is within 30M, which is much less than GPT2-Small with 110M parameters. The GPT-2 variants details are presented in Table \ref{gpt2-version} in Appendix \ref{apdx:vit-gpt-details} and the ones for the \textsc{GPTs-1K} dataset's generation can be found in Appendix \ref{appendix_gpts1k}.

Importantly, these datasets are smaller than Large Language Models (LLMs) and similarly large vision models; this is by design. The purpose is to reduce the computation required for learning to predict parameters while giving a continuous range of scale (from tiny to large) so that \textsc{LoGAH} can generalize to large models after training.

\begin{table}
\caption{Details of \textsc{LoGAH} variants and \textsc{GHN-3} variants. All \textsc{LoGAH} variants are set with $K=2048\cdot 16$. We estimate the train time of each model based on meta-batch $m=1$ and the CIFAR-100 dataset for 300 epochs.}
\label{ghn_lora_versions}
\begin{center}
\begin{tabular}{lcccccccc}
\toprule
\textbf{Model} & \textit{r} & \textit{L} &\textit{d} & \textit{H}& Max Width  &P & Train Time \\
\midrule
LoGAH-Tiny & 32  & 3 & 64 & 8 & 2048 & 2.5M & 7.05h\\
LoGAH-Small & 90 & 5 & 128 &16 & 2048& 21.4M & 7.25h\\
LoGAH-Base & 128 & 5 & 256 &16 & 2048 &78.2M & 10.30h\\
LoGAH-Large & 256 & 12 & 256 & 16 & 2048&289.4M & 21.0h\\ 
\midrule
GHN-3-Tiny & - & 3 & 64 & 8 & 64& 6.9M & 7.20h\\
GHN-3-Small & - & 5 & 128 & 16 & 128 & 35.8M& 7.75h\\
GHN-3-Large & - & 12 & 256 & 16& 256 & 214.7M & 12.40h\\
GHN-3-XLarge & - & 24 & 384 & 16 &384 & 654.4M& 24.0h\\
\bottomrule
\end{tabular}
\end{center}
\end{table}


\section{Experiments}
\label{experiments}
We evaluate if networks (i.e. ViT and GPT-2) initialized with the parameters $\textbf{w}_{\text{pred}}$ predicted by \textsc{LoGAH} can perform better than those by GHN-3 and random initialization after fine-tuning. 

\textbf{\textsc{LoGAH} Variants.} We provide four different scales of \textsc{LoGAH} from \textsc{Tiny} to \textsc{Large}, by gradually increasing the number of layers $L$, hidden size $d$, heads $H$, as well as the low-rank $r$. We also compare the number of parameters and estimate the training time difference between \textsc{LoGAH} with GHN-3, shown in Table \ref{ghn_lora_versions}. We highlight that GHN-3 and \textsc{LoGAH} are trained only once on each dataset, so that the same model can predict parameters for many architectures making the training cost of GHN-3 and \textsc{LoGAH} amortized.

\textbf{\textsc{GHN} Training Setup.} The GHN models, including GHN-3 and our \textsc{LoGAH}, are trained for $300$ epochs on \textsc{ViTs-1K} and \textsc{GPTs-1K} datasets. For ViT, we conduct experiments on the following datasets: CIFAR-10, CIFAR-100 \citep{krizhevsky2009learning} (with batch size $b=64$) and ILSVRC-2012 ImageNet \citep{russakovsky2015imagenet} (with batch size $b=128$). We train the models using automatic mixed precision in PyTorch with a cosine annealing learning rate schedule starting at $lr=3e-4$, weight decay $\lambda=1e-2$, and predicted parameter regularization $\gamma=3e-5$~\citep{knyazev2023scale}. For GPT-2 experiments we use the WikiText dataset \citep{merity2016pointer}, and use $lr=1e-4$ with batch size $b=6$, while keeping the other hyperparameters as before. All GHN models, including GHN-3 and \textsc{LoGAH}, are trained separately on each task dataset.

\begin{table}
\caption{CIFAR-10, CIFAR-100 and ImageNet top-1 accuracy (\%) on ViT-Small and ViT-Base in different initialization settings. ViT models on CIFAR datasets are fine-tuned for 100 epochs in each initialization setting, while for ImageNet dataset, ViTs are fine-tuned for 30 epochs due to the time cost consideration. All GHN models can be trained on a single NVIDIA 4090 GPU when meta-batch $m=1$ with /m1 suffix. On ImageNet, we only train GHN-3-Tiny.}
\label{vit_experiments1}
\begin{center}

\begin{tabular}{lcccccc}
\toprule
\multicolumn{1}{c}{\textbf{Initialization}}&
\multicolumn{2}{|c|}{CIFAR-10}&
\multicolumn{2}{|c}{CIFAR-100}&
\multicolumn{2}{|c}{ImageNet}\\
 & ViT-Small & ViT-Base & ViT-Small & ViT-Base & ViT-Small & ViT-Base\\
\midrule
\textsc{RandInit} & 83.93 & 84.74& 58.73& 57.00 & 62.15 &63.61\\
\textsc{OrthInit} & 80.11 & 84.24 & 58.97 & 57.48 &63.15 &63.86 \\
\midrule
GHN-3-T/m1 &81.29 & 78.18 &56.33 &56.70 & 38.79 & 31.89\\
GHN-3-S/m1 &83.61 & 82.57&57.37 &59.93 & - & -\\
GHN-3-L/m1 &84.69 &82.84 &57.48 & 56.57 & - & -\\
\midrule
LoGAH-T/m1 & 82.87 &82.87  &60.05 & 56.77 & 62.16 & 61.68 \\
LoGAH-S/m1 & \textbf{86.09} & \textbf{86.35} &61.05 &62.38 & 62.49 & 63.29\\
LoGAH-B/m1 &85.16  &85.39 &60.58 & 60.02 &59.00  & 61.22\\
LoGAH-L/m1 & 83.22 &84.40 &\textbf{62.19} &\textbf{63.47} & 63.00 & 63.70 \\
\bottomrule
\end{tabular}
\end{center}
\end{table}

\subsection{ViT Experiments.}
\subsubsection{Overall Comparision on CIFAR-10, CIFAR-100 and ImageNet}
We test ViT-small and ViT-base on CIFAR-10, CIFAR-100 \citep{krizhevsky2009learning} and ILSVRC-2012 ImageNet \citep{russakovsky2015imagenet} shown in Table \ref{vit_experiments1} with different initialization methods: (1) random initialization (\textsc{RandInit}) implemented by default in PyTorch, (2) orthogonal initialization (\textsc{OrthInit}) \citep{saxe2014exact}, (3) parameters predicted by GHN-3, and (4) parameters predicted by \textsc{LoGAH}. From Table \ref{vit_experiments1}, we observe that \textsc{LoGAH} generally outperforms \textsc{RandInit}, \textsc{OrthInit} and GHN-3.

\begin{wrapfigure}{r}{0.48\textwidth}
  \centering
{\includegraphics[width=0.48\textwidth]{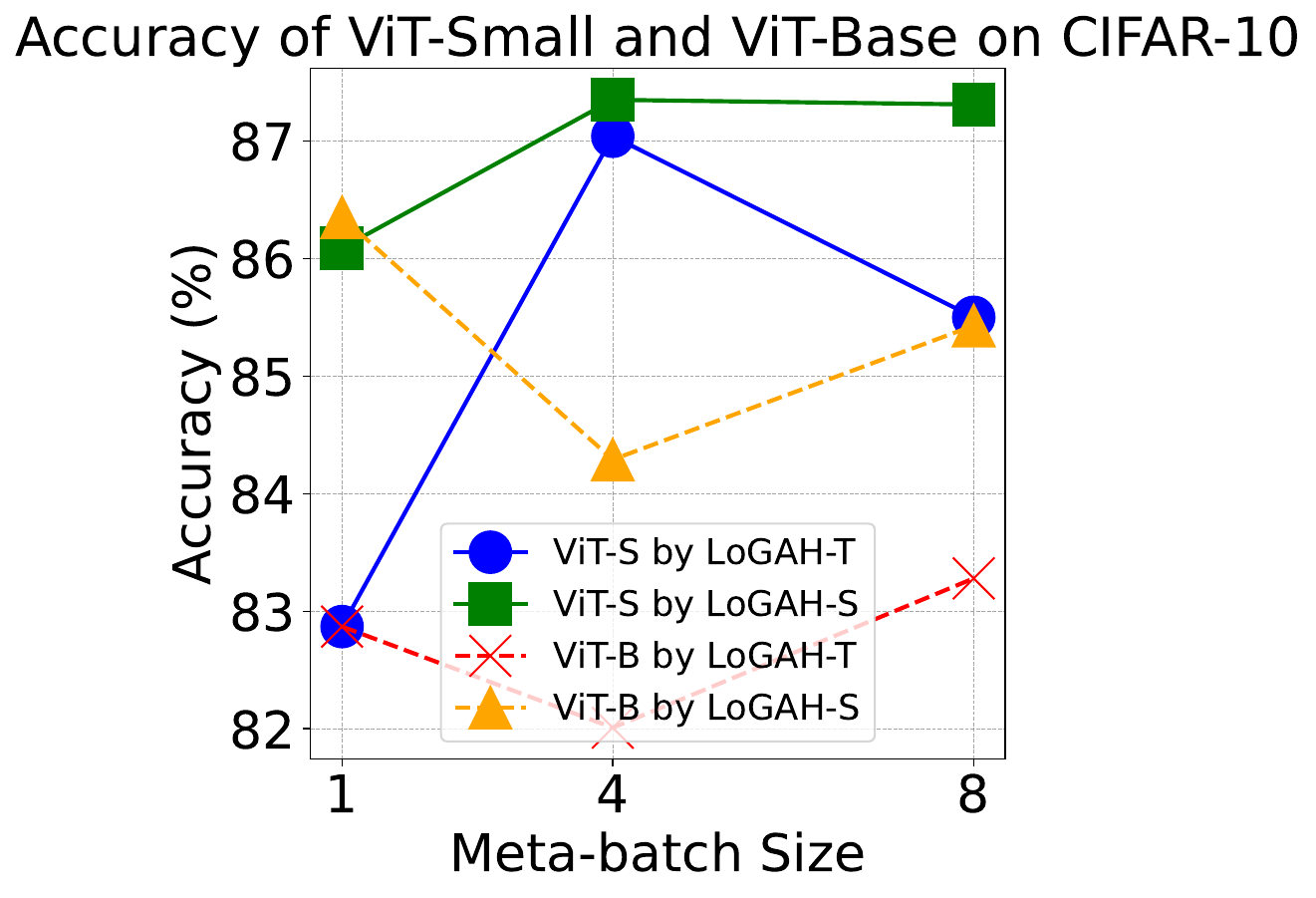}}
  \caption{CIFAR-10 top-1 accuracy (\%) on ViT-Small and ViT-Base where \textsc{LoGAH} is trained with different meta-batch size $m$.}
\label{ViT-GHN-metabatch}
\end{wrapfigure}

The ViT models are fine-tuned for 100 epochs on CIFAR, and 30 epochs on ImageNet datasets. For ViT-Small, we set the learning rate range as $\{0.1, 0.2, 0.3, 0.4, 0.5\}$. For ViT-Base and ViT-Large, we set it as $\{0.03, 0.04, 0.05, 0.06, 0.1\}$. All ViT models are fine-tuned using SGD with the same setting as \citet{knyazev2023scale}, but with weight decay $1e-2$. We report the best validation accuracy among all learning rates. 

\textbf{CIFAR-10.} \textsc{LoGAH-Small} presents the best performance (i.e. 2.16\% improvement than \textsc{RandInit}, and 1.4\% improvement than \textsc{GHN-3-L} in ViT-small) with only $21.41$M parameters, compared to \textsc{GHN-3-L} with $214.7$M parameters, $10$x larger than \textsc{LoGAH-Small}, which demonstrates the effectiveness of our low-rank decoder. For ViT-base models, \textsc{GHN-3-L} has the closest performance to our smallest version of our models: \textsc{LoGAH-Tiny}.

\textbf{CIFAR-100.} \textsc{LoGAH-Large} outperforms other models with a big improvement, especially compared to \textsc{RandInit} (3.46\% on ViT-Small and 6.47\% on ViT-Base), while GHN-3's performances are even worse than the random initializations on ViT-Small.

\textbf{ImageNet.} Considering the time cost for training GHNs on ImageNet, for GHN-3, we only train GHN-3-T/m1 in this dataset for comparison. The random initialization methods, including \textsc{RandInit} and \textsc{OrthInit}, show comparable performances with \textsc{LoGAH}. In this dataset, we do not observe a big improvement as in CIFAR-10 and CIFAR-100. However, \textsc{LoGAH-T} performs much better than GHN-3-T (62.16 vs. 38.79 for ViT-Small).

\subsubsection{Effect of meta-batch size $m$ on \textsc{LoGAH}}

In this section, we study the effect of the meta-batch size $m$ of \textsc{LoGAH-Tiny} and \textsc{LoGAH-Small} on ViT-Small, ViT-Base and ViT-Large (shown in Figure \ref{ViT-GHN-metabatch} for CIFAR-10 and Table \ref{meta-batch-effect2} for CIFAR-100). We increase the value of $m$ from $1$ to $4, 8$ and train \textsc{LoGAH} respectively.

\textbf{CIFAR-10.} In Figure \ref{ViT-GHN-metabatch}, we can notice that increasing the meta-batch size $m$ can significantly improve the ViT-Small's performance (i.e. from $82.87$ to $87.04$ by \textsc{LoGAH-Tiny}, and from $86.09$ to $87.35$ by \textsc{LoGAH-Small} when setting $m=4$), which is even better than larger \textsc{LoGAH} models trained with $m=1$. However, a reverse pattern is observed in ViT-Base, increasing $m$ to $4$ will worsen the performance on CIFAR-10.
\begin{wraptable}{r}{0.53\textwidth}
\caption{CIFAR-100 top-1 accuracy (\%) on ViT-Base and ViT-Large where $m=4, 8$ for training \textsc{LoGAH}.}
\label{meta-batch-effect2}
\begin{center}
\begin{tabular}{lcc}
\toprule
\textbf{Initialization} & ViT-Base & ViT-Large\\
\midrule
\textsc{RandInit} & 57.00 & 55.62 \\
\midrule
LoGAH-T/m1 & 56.77 & 55.07 \\
\rowcolor{Apricot}
LoGAH-T/m4 &\underline{60.61} &\underline{59.88} \\
LoGAH-T/m8 &51.12 & 50.39\\
\midrule
\rowcolor{SpringGreen}
LoGAH-S/m1 & \textbf{62.38} & \textbf{59.90}\\
LoGAH-S/m4 & 55.00 & 54.62\\
LoGAH-S/m8 & 51.58 & 51.61 \\
\bottomrule
\end{tabular}
\end{center}
\vspace{-10pt}
\end{wraptable}

\textbf{CIFAR-100.} For the CIFAR-100 dataset, we extend to the larger model: ViT-Large. Enlarging the meta-batch size properly can significantly improve the small \textsc{LoGAH}'s performance and help it achieve similar results as the larger one (e.g. \colorbox{Apricot}{\textsc{LoGAH-T/m4}} vs. \colorbox{SpringGreen}{\textsc{LoGAH-S/m1}}). Another interesting finding is that when setting the meta-batch size $m=8$, both \textsc{LoGAH}'s performance drops dramatically. 
One potential reason might lie in the smaller model sizes in \textsc{ViTs-1K}, and larger meta-batch may cause the overfitting problem.

\subsection{GPT-2 Experiments}
\label{gpt2-experiments}

Considering the time and other resource costs\footnote{For example, we did not train on the OpenWebText dataset \citep{Gokaslan2019OpenWeb}, since it is too large.
}, we only apply two smallest \textsc{LoGAH}, \textsc{LoGAH-T/m2} and \textsc{LoGAH-S/m2}, in the GPT-2 experiments. This section investigates the Causal Language Modeling (CLM) task. We test the model's performance on the WikiText dataset\citep{merity2016pointer}. In detail, we choose GPT-2-Small and GPT-2-Medium for the \verb|wikitext-2-raw-v1| dataset. For the \verb|wikitext-103-raw-v1| dataset, we select GPT-2-Medium and GPT-2-Large. All models are trained with the randomly-initialized parameters (i.e. \textsc{RandInit}), which is implemented by default in HuggingFace \citep{wolf2019huggingface}.

\textbf{GPT-2 Training Setup.} All the GPT-2 models are fine-tuned for 100 epochs on each dataset. We use DeepSpeed \citep{rajbhandari2020zero} for improved training efficiency with GPT-2-Medium and GPT-2-Large on WikiText-103. With 6$\times$ NVIDIA 4090 GPUs, we train the models by AdamW with learning rate as $3e-6$, weight decay as $1e-2$, batch size as $4$ for GPT-2-Medium and as $2$ for GPT-2-Large.

\textbf{Results.} Table \ref{gpt_experiments1} shows the results. The performance is improved more significantly on larger GPT-2 models and larger datasets (e.g. training GPT-2-Large on the WikiText-103 dataset), which demonstrates that even our smallest model (2.5M) can predict much better parameters than a random initialization for a large model (774M).

\begin{table}
\caption{Perplexity score of the GPT-2 experiments.}
\label{gpt_experiments1}
\begin{center}
\begin{tabular}{lcccc}
\toprule
\multicolumn{1}{c}{\textbf{Initialization}}&
\multicolumn{2}{|c|}{WikiText-2}&
\multicolumn{2}{|c}{WikiText-103}\\
  & GPT-2-Small & GPT-2-Medium & GPT-2-Medium & GPT-2-Large \\
\midrule
\textsc{RandInit} & 250 & 350 & 22.32& 32.41\\
\midrule
LoGAH-T/m2 &\textbf{227} & \textbf{219} & \textbf{18.79} & 27.18 \\
LoGAH-S/m2 &238 &284 & 19.89 & \textbf{24.08}\\
\bottomrule
\end{tabular}
\end{center}
\end{table}

\subsection{Qualitative Analysis}
We analyze the diversity of predicted parameters following experiments in \citet{knyazev2023scale, knyazev2021parameter}. In detail, we predict parameters for ViT and GPT-2, and collect one or two frequently occurring tensor shapes in each model. Then we compute the absolute cosine distance between all pairs of parameter tensors of the same shape and average it (Table \ref{diversity_exps}). \textsc{LoGAH} predicts more diverse parameters than GHN-3 in general on ViT models, especially in ViT-Small on CIFAR-100, which is also consistent with the better performance in Table \ref{vit_experiments1}. Another interesting finding is that \textsc{LoGAH-Tiny} is better at predicting more diverse square parameters (e.g. (768,768) and (1024,1024)) than \textsc{LoGAH-Small}.

\begin{table}
\caption{Diversity of the parameters predicted by GHN-3, \textsc{LoGAH} vs. Pretrained ($*$: or trained by SGD from \textsc{RandInit} if pretrained parameters are unavailable) measured on the ViT and GPT-2. \textcolor{gray}{Pretrain*}: ViT-Small is trained by SGD from \textsc{RandInit} for 100 epochs on CIFAR-100; for ViT-Base and ViT-Large we use the parameters pretrained on ImageNet; for GPT-2-Medium, we also load the available pretrained parameters in HuggingFace \citep{wolf2019huggingface}. }
\label{diversity_exps}
\begin{center}
\begin{small}
\begin{tabular}{lccccccc}
\toprule
\multicolumn{1}{c}{\textbf{Method}}&
\multicolumn{6}{c}{\textbf{Parameter Tensor Shape}}\\
\multicolumn{1}{c|}{}&
\multicolumn{1}{|c|}{CIFAR-100}&
\multicolumn{1}{|c|}{ImageNet}&
\multicolumn{2}{|c|}{ImageNet}&
\multicolumn{2}{|c}{WikiText-103}\\
\multicolumn{1}{c|}{}&
\multicolumn{1}{|c|}{ViT-S}&
\multicolumn{1}{|c|}{ViT-B}&
\multicolumn{2}{|c|}{ViT-L}&
\multicolumn{2}{|c}{GPT-2-M}\\
& (1536,384) & (768,768)&(1024,1024)&(3072,1024)& (1024,1024) & (1024,4096)\\
\midrule
\textcolor{gray}{Pretrain$^*$} &\textcolor{gray}{0.647} &\textcolor{gray}{0.747}&\textcolor{gray}{0.878}&\textcolor{gray}{0.842} &\textcolor{gray}{0.839} &\textcolor{gray}{0.894}\\
\midrule
GHN-3-T&0.191&0.513&0.402&0.420& - &-\\
GHN-3-S&0.290&-&-&-&-&-\\
\midrule
\textsc{LoGAH-T}&0.284&\textbf{0.563}&\textbf{0.410}&0.385 & \textbf{0.420} & 0.393\\
\textsc{LoGAH-S}&\textbf{0.442}&0.413&0.232&0.329 &0.400 &\textbf{ 0.496}\\
\bottomrule
\end{tabular}
\end{small}
\end{center}
\end{table}

\subsection{Transfer Learning Experiments}
\begin{wrapfigure}{r}{0.48\textwidth}
\vspace{-15pt} 
  \centering
{\includegraphics[width=0.48\textwidth]{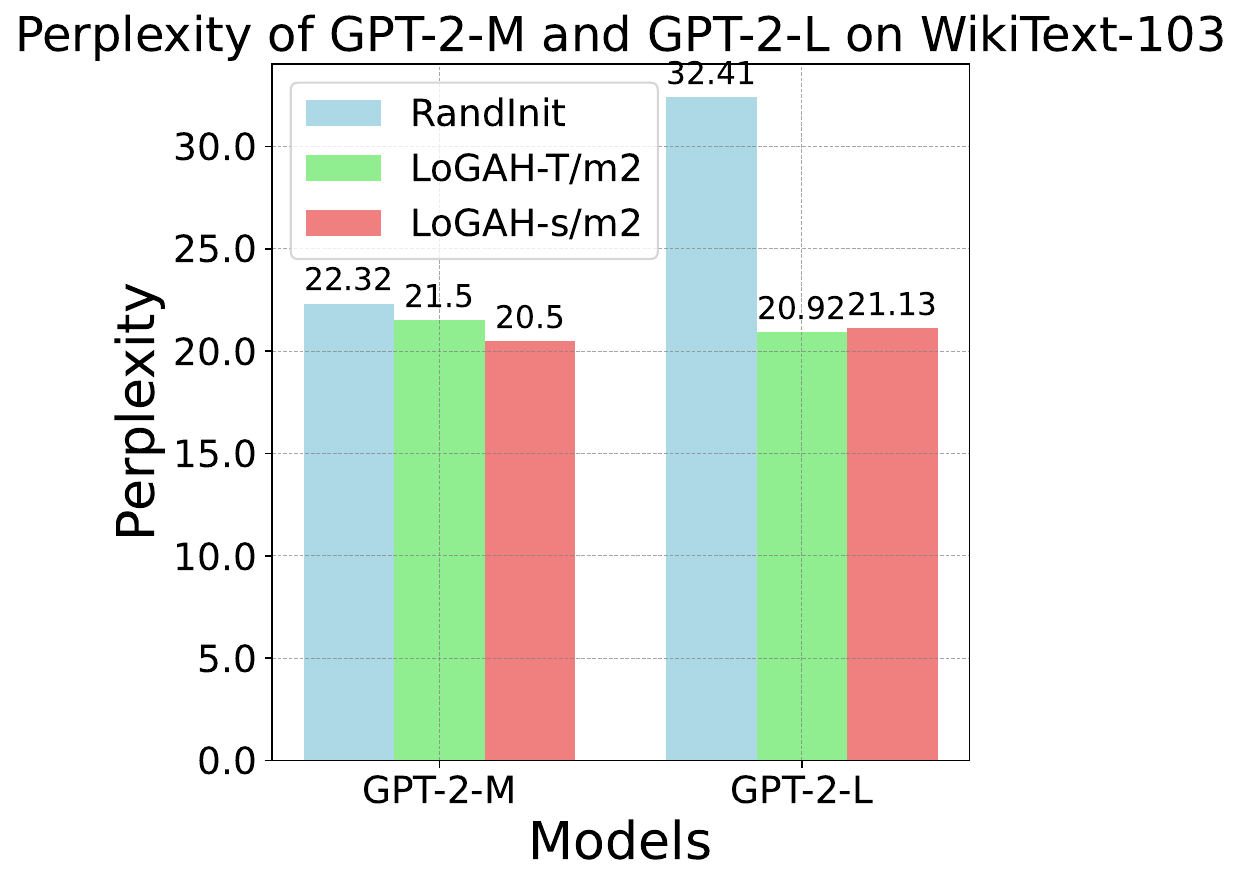}}
  \caption{GPT-2 transfer learning experiments. \textsc{LoGAH} are trained on WikiText-2 and GPT-2 models are fine-tuned on WikiText-103 based on \textsc{LoGAH}'s predicted parameters.}
\label{gpt-transfer-exp}
\vspace{-20pt} 
\end{wrapfigure}
In this section, we explore the setting when \textsc{LoGAH} is trained on one dataset, but is used to produce a parameter initialization for another (potentially more difficult) dataset. For the ViT experiments, we conduct the transfer learning experiments from CIFAR-10 to CIFAR-100, and from CIFAR-100 to ImageNet. For the GPT-2 experiments, we consider experiments from WikiText-2 to WikiText-103.

\begin{figure}
  \centering
  \centerline{\includegraphics[width=0.9\columnwidth]{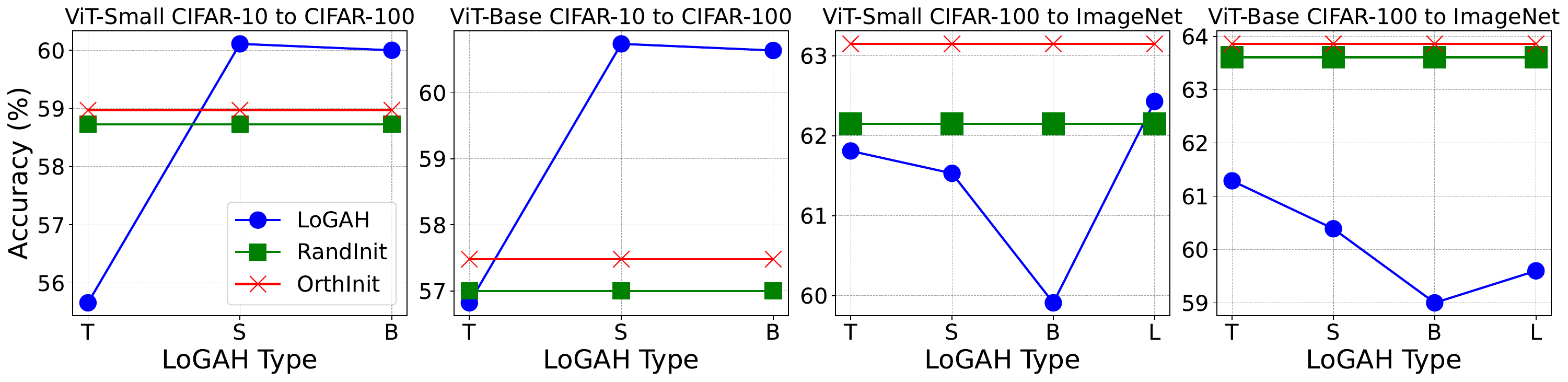}}
  \caption{ViT transfer learning experiments. We use \textsc{LoGAH} trained on CIFAR-10 (\textit{resp.} CIFAR-100) to predict ViT's parameters, then ViT is trained on CIFAR-100 (\textit{resp.} ImageNet). T, S, B and L denotes \textsc{Tiny, Small, Base} and \textsc{Large} versions of \textsc{LoGAH} respectively.}.
  \vspace{-10pt}
  \label{transfer-vit}
\end{figure}

\textbf{ViT Experiments.} 
In detail, we re-initialize the classification layer \citep{knyazev2023scale} using a Kaiming normal distribution \citep{he2015delving} with 100 and 1,000 outputs, for transferring to CIFAR-100 and ImageNet respectively, then we fine-tune the entire network. The results are shown in Figure \ref{transfer-vit}. On CIFAR-100, ViTs initialized by the \textsc{LoGAH-S, LoGAH-B} outperform random initialization. It is still a demanding task for \textsc{LoGAH} to work on the ImageNet dataset, and \textsc{LoGAH-T} is generally better than larger ones, which suggests that parameters that work well on CIFAR-100 may not be optimal for ImageNet. However, compared to GHN-3-T in Table \ref{vit_experiments1}, it shows superior performances in both ViT-Small and ViT-Base.

\textbf{GPT-2 Experiments.} 
We keep the same setting as in Section \ref{gpt2-experiments}, and fine-tune GPT-2-Medium and GPT-2-Large on WikiText-103 loaded with \textsc{LoGAH} predicted parameters, which are trained on the WikiText-2 dataset. According to Figure \ref{gpt-transfer-exp}, \textsc{LoGAH}'s predicted parameters are also a good initialization for fine-tuning GPT-2 models on WikiText-103, especially for GPT-2-Large. 

According to the above results, the parameters predicted by \textsc{LoGAH} present a desired transfer learning ability from one easier to another harder task. This improvements are more significant if the data distribution is close (e.g. from CIFAR-10 to CIFAR-100). Moreover, this property can help reduce the training time for \textsc{LoGAH} to predict good parameters, where we do not need to train \textsc{LoGAH} on a large-scale dataset.

\section{Related Work}
\vspace{-5pt}
\textbf{Large Models Pretraining.} The large-scale pretrained models first appeared in the NLP field \citep{yin2022survey, guo2022threats}. The improvement and success are mainly attributed to self-supervised learning and Transformer \citep{vaswani2023attention}. More and more large language models are developed based on it, extending to larger sizes for better performance under pretraining with massive data \citep{devlin-etal-2019-bert, brown2020language, touvron2023llama}. Inspired by the advancement of Transformer, many Transformer-based vision models are also proposed, and some pretraining methods have been explored  \citep{dosovitskiy2021image, carion2020endtoend, he2021masked,pmlr-v119-chen20s}. Our work focuses on predicting parameters for two Transformer-based models (ViT and GPT-2) to reduce pretraining costs.

\textbf{Parameter Prediction.} Hypernetworks \citep{ha2016hypernetworks} are often leveraged for predicting model's parameter. Many research works have extended the hypernetwork's capability to generalize on unseen architectures \citep{zhang2018graph, Nirkin_2021_CVPR, knyazev2021parameter}, datasets \citep{requeima2020fast,10.1145/3442381.3449908, zhmoginov2022hypertransformer, kirsch2024generalpurpose}, or to generate interpretable networks \citep{liao2023generating}. Our paper is also based on Graph HyperNetworks (GHNs), but overcomes the extreme increase of parameters needed in previous GHNs. \textsc{LoGAH} can support larger models with just $1\%$ parameters, showing a better ability to predict parameters for larger networks.

\textbf{Initialization and Learning to Grow Models.} Several methods have improved on random initialization by learning from data~\citep{dauphin2019metainit,yang2022towards}. However, GHN-3~\citep{knyazev2023scale} showed better performance making it a favourable approach to build on. Other methods learn to initialize a bigger model from a smaller pretrained model~\citep{evci2022gradmax,wang2023learning}. These methods reduce training time, however, a smaller pretrained model of exactly the same architecture as the target model is not always available, which limits the approach.

\section{Limitations}
\vspace{-5pt}
Although our model \textsc{LoGAH} shows outstanding performances compared to GHN-3 and other random initialization methods across the extensive experiments, there are still limitations. 
Most importantly, due to the consideration of time and resource costs, we conduct GPT-2 experiments only on the WikiText dataset and only with our two smallest models. 
Furthermore, in order to predict parameters for drastically novel architectures (e.g.~\citep{gu2023mamba}), the GHN might be needed to be trained to avoid a big distribution shift.
In future work, it would be intriguing to show \textsc{LoGAH}'s ability on modern LLMs~\citep{touvron2023llama}. 


\section{Conclusion}
In this work, we propose \textsc{LoGAH}, a low-rank Graph HyperNetwork (GHN) that addresses two issues of previous GHN-3. First, the low-rank decoder avoids copying small chunks of parameters multiple times when predicting a large shape parameter. Second, contrary to GHN-3, it does not require an exceptional number of trainable parameters to support wider or larger models, and our smallest \textsc{LoGAH} is only about 2.5M. We conduct extensive experiments on two representative transformer-based models (ViT in vision and GPT-2 in language) to show superior efficacy compared to GHN-3 and random initialization methods. Furthermore, the generalization ability of \textsc{LoGAH} from a simple to another more difficult dataset is also verified in the transfer learning experiments. 




\bibliographystyle{unsrtnat}
\bibliography{main}

\appendix
\section{Details of the amounts of parameters of decoders in GHN-3}
\label{apdx: ghn3-decoder}
The theory amount of parameters of decoders in GHN-3 is shown below:
\begin{align}
\label{eq:ghn3-dec-params}
    4\times \text{in\_feature}\times d\times h\times w + \text{MLP\_d}_1\times \text{MLP\_d}_2 + \text{MLP\_d}_2\times d^2 + d\times \text{num\_class}
\end{align}

where $\text{in\_feature}$ is the input feature's dimension of the decoder (set as $d$ in GHN-3), and $\text{MLP\_d}_1, \text{MLP\_d}_2$ denote the dimension of $1_{\text{st}}$ and $2_{\text{nd}}$ layers of MLP (set as $4d$ and $8d$ in experiments respectively), $h, w$ are the last two dimensions of the predicted tensor's shape (set as $16$) and $\text{num\_class}$ is the number of classes of the dataset. Thereby, we can simplify Equation \eqref{eq:ghn3-dec-params} to \eqref{num_param_decoder1}.

\section{Details of MLPs in the decoder of \textsc{LoGAH}}
\label{mlps_ghn_lora}
The MLPs has 4 layers and the activation function $\sigma(\cdot)$ is ReLU \cite{fukushima1975cognitron}:
\begin{align}
  \textbf{x} &= M_3\Bigg(\sigma\bigg(M_2\Big(\sigma\big(M_1(\textbf{H}) \big) \Big) \bigg) \Bigg)\\
  \textbf{x} &= \text{reshape}(\textbf{x})\in\mathbb{R}^{|V|\times 2r\times r}\\
  \textbf{x} &=\text{reshape}\big(M_4(\sigma(\textbf{x}))\big)\in\mathbb{R}^{|V|\times 2K\times r}
\end{align}
where $M_i, i\in \{1,2,3,4\}$ are learnable matrices:
\begin{align*}
    M_1&\in\mathbb{R}^{d\times 4d}, M_2\in\mathbb{R}^{4d\times 8d}\\
    M_3&\in\mathbb{R}^{8d\times 2r^2}, M_4\in\mathbb{R}^{r\times K}
\end{align*}

We also provide the code implementation of it as shown in Figure \ref{code:low-rank-decoder-logah}.

\lstset{frame=tb,
  language=Python,
  aboveskip=3mm,
  belowskip=3mm,
  showstringspaces=false,
  columns=flexible,
  basicstyle={\small\ttfamily},
  breaklines=true,
  breakatwhitespace=true,
  tabsize=3,
  backgroundcolor=\color{backcolour},   
    commentstyle=\color{codegreen},
    keywordstyle=\color{magenta},
    numberstyle=\tiny\color{codegray},
    stringstyle=\color{codepurple},
}
\begin{figure}[]
    \centering
    \begin{lstlisting}
class ConvDecoder3LoRA(nn.Module):
    def __init__(self,
                 in_features,
                 ck=32,
                 r=32,
                 hid=(64,),
                 num_classes=None):
        super(ConvDecoder3LoRA, self).__init__()

        assert len(hid) > 0, hid
        self.r = r
        self.ck = ck
        self.num_classes = num_classes
        self.mlp = MLP(in_features=in_features,
                       hid=(*hid, r*2*r), 
                       activation='relu',
                       last_activation=None)

        self.l2 = nn.Linear(int(r), ck)
        self.relu = nn.ReLU(inplace=True)
        
        self.seq = nn.Sequential(
            self.relu,
            self.l2
        )


    def forward(self, x, max_shape=(1,1,1,1), class_pred=False, n_dim = 4):
        if class_pred:
            n_dim = 2
        x = self.mlp(x).view(-1, 2*self.r, self.r) # [b, 2*r, r]
        x = self.seq(x).view(-1, 2*self.ck, self.r) # [b, 2*ck, r]
        A, B_t = torch.split(x, self.ck, dim=1) # A=[b, ck, r] and B=[b, ck, r]
        B = B_t.transpose(1,2) # A=[b, ck, r] and B=[b, r, ck]
        # fix shape of A and B before matmul through indexing
        c_out, c_in, k_out, k_in = max_shape
        A = A[:, :(c_out*k_out), :] # [b, c_out*k_out, r]
        B = B[:, :, :(c_in*k_in)] # [b, r, c_in*k_in]
        W = torch.bmm(A, B) # [b, c_out*k_out, c_in*k_in]
        if n_dim == 1: # We want [c_out]
            assert c_in == 1 and k_out == 1 and k_in == 1
            W = W.reshape(-1, c_out)
        elif n_dim == 2: # we already have a 2D matrix
            pass
        elif n_dim == 4:
            W = W.reshape(-1, c_out, k_out, c_in, k_in).transpose(2, 3) # [b, c_out, c_in, k_out, k_in]
        else:
            raise NotImplementedError("n_dim must be 1 or 2 or 3")
        #print(W.shape)
        return W
    \end{lstlisting}
     \caption{Code for Low-rank decoder in \textsc{LoGAH}.}
     \label{code:low-rank-decoder-logah}
\end{figure}

\lstset{frame=tb,
  language=Python,
  aboveskip=3mm,
  belowskip=3mm,
  showstringspaces=false,
  columns=flexible,
  basicstyle={\small\ttfamily},
  breaklines=true,
  breakatwhitespace=true,
  tabsize=3,
  backgroundcolor=\color{backcolour},   
    commentstyle=\color{codegreen},
    keywordstyle=\color{magenta},
    numberstyle=\tiny\color{codegray},
    stringstyle=\color{codepurple},
}
\begin{figure}[]
    \centering
    \begin{lstlisting}
    layers = np.random.randint(3, 10)
    if layers > 5:
        dim_min = 128
        dim_max = 256
    elif layers > 3:
        dim_min = 256
        dim_max = 384
    else:
        dim_min = 384
        dim_max = 512
    
    hidden_dim = np.random.choice(np.arange(dim_min, dim_max+1, 32))
    mlp_dim = hidden_dim * 4

    if hidden_dim % 12 == 0:
        heads = np.random.choice([3, 6, 12])
    elif hidden_dim % 6 == 0:
        heads = np.random.choice([3, 6])
    elif hidden_dim % 3 == 0:
        heads = 3
    else:
        heads = np.random.choice([4, 8])
    
    
    
    net = _vision_transformer(
        patch_size = 2,
        num_layers = layers,
        num_heads = heads,
        hidden_dim = hidden_dim,
        mlp_dim = mlp_dim,
        num_classes = 100,
        image_size = 32,
        weights = None,
        progress = False,
    )
    
    \end{lstlisting}
     \caption{Code for generating ViT-style models used for \textsc{ViTs-1K} dataset.}
     \label{code:vit-1k}
\end{figure}

\lstset{frame=tb,
  language=Python,
  aboveskip=3mm,
  belowskip=3mm,
  showstringspaces=false,
  columns=flexible,
  basicstyle={\small\ttfamily},
  breaklines=true,
  breakatwhitespace=true,
  tabsize=3,
  backgroundcolor=\color{backcolour},   
    commentstyle=\color{codegreen},
    keywordstyle=\color{magenta},
    numberstyle=\tiny\color{codegray},
    stringstyle=\color{codepurple},
}
\begin{figure}[]
    \centering
    \begin{lstlisting}
    n_layer = np.random.randint(3, 10)

    if n_layer > 5:
        dim_min = 72
        dim_max = 176
    elif n_layer > 3:
        dim_min = 128
        dim_max = 176
    else:
        dim_min = 176
        dim_max = 256
    
    n_embd = np.random.choice(np.arange(dim_min, dim_max+1, 8))

    if n_embd % 8 == 0:
        n_head = 8
    elif n_embd % 6 == 0:
        n_head = 6
    elif n_embd % 4 == 0:
        n_head = 4

    config = GPT2Config(
        bos_token_id=tokenizer.bos_token_id,
        eos_token_id=tokenizer.eos_token_id,
        n_embd=int(n_embd),
        n_layer=int(n_layer),
        n_head=int(n_head),
        tie_word_embeddings=False,
    )
    model = GPT2LMHeadModel(config)
    
    \end{lstlisting}
     \caption{Code for generating GPT-2-style models used for \textsc{GPTs-1K} dataset.}
     \label{code:gpt-1k}
\end{figure}

\section{Details of variants of ViT and GPT-2}
\label{apdx:vit-gpt-details}
We provide the details of ViT and GPT-2 in different sizes. $L, D, H, P$ denotes the numbers of layers, heads, hidden size and parameters, respectively.

\begin{minipage}{.5\textwidth}
\centering
\begin{small}
\begin{tabular}{lcccccc}
\toprule
Model &  \textit{L} &\textit{D} & MLP size & \textit{H}& P \\
\midrule
ViT-S & 12 & 384 & 1536 & 6 & 22M\\
ViT-B& 12 & 768 & 3072 & 12 & 86M\\
ViT-L & 24 & 1024 & 4096 & 16 & 307M\\
\bottomrule
\end{tabular}
\captionof{table}{Details of ViT variants}
\label{vit-version}
\end{small}
\end{minipage}
\begin{minipage}{.5\textwidth}
\centering
\begin{small}
\begin{tabular}{lccccc}
\toprule
Model &  \textit{L} &\textit{D} & \textit{H}& P \\
\midrule
GPT-2-S & 12 & 768 &12 & 110M\\
GPT-2-M & 24 &1024 & 16 & 345M\\
GPT-2-L & 36 & 1280 & 20 &774M\\
\bottomrule
\end{tabular}
\captionof{table}{Details of GPT-2 variants}
\label{gpt2-version}
\end{small}
\end{minipage}

\section{Distribution of \textsc{ViTs-1K} and \textsc{GPTs-1K} datasets}
\begin{figure}[!tbp]
  \centering
  \subfloat[The parameters distribution in \textsc{ViTs-1K}.]{\includegraphics[width=0.5\textwidth]{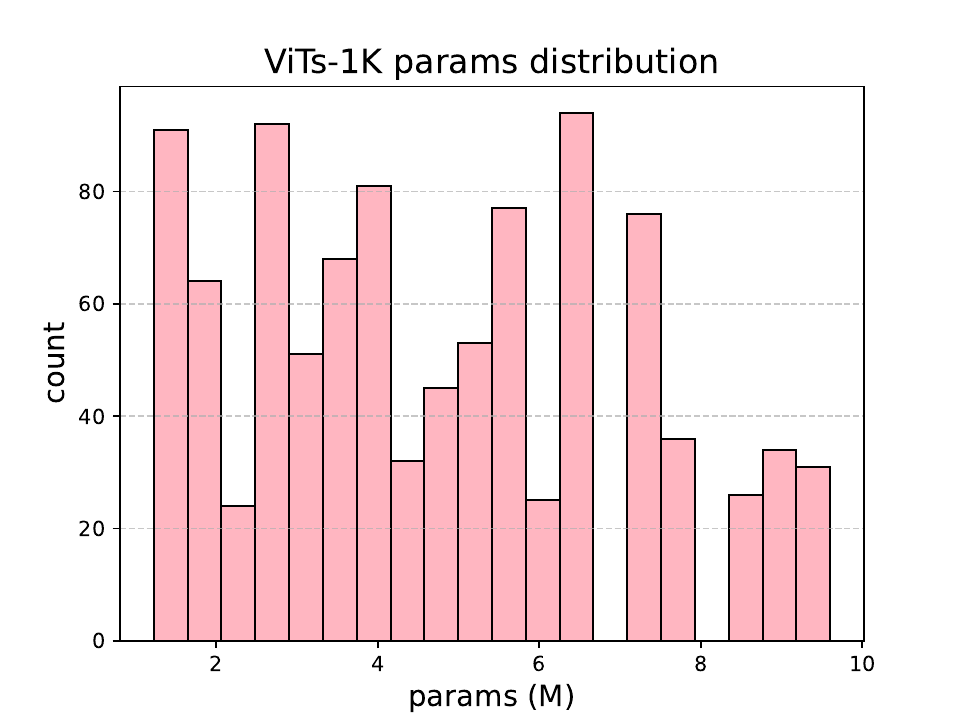}\label{vits1k-params-dis}}
  \hfill
  \subfloat[The parameters distribution in \textsc{GPTs-1K}.]{\includegraphics[width=0.5\textwidth]{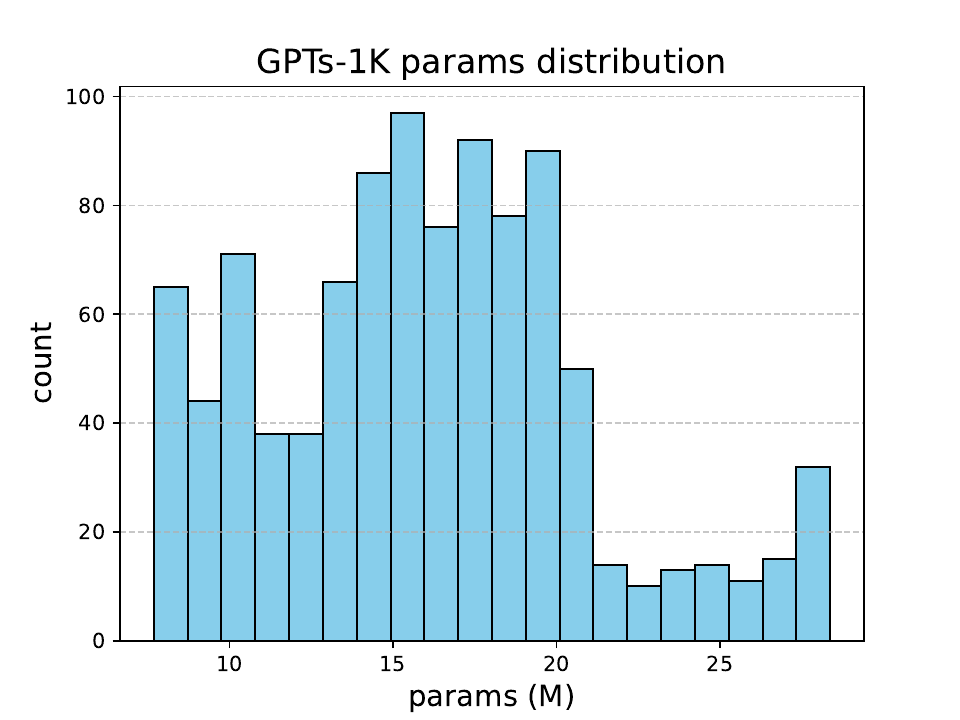}\label{gpt2s1k-params-dis}}
  \caption{The parameters distribution of \textsc{ViTs-1K} and \textsc{GPTs-1K} datasets}
  \label{vit1k-gpt1k-dis}
\end{figure}

The distributions of \textsc{ViTs-1K} and \textsc{GPTs-1K} datasets are shown in Figure \ref{vit1k-gpt1k-dis}.

\section{Details of generating \textsc{ViTs-1K} dataset}
\label{appendix_vits1k}
As mentioned above, we change the values in layers $L$, heads $H$, and hidden size $D$ of ViT, as well as restricting these models size. The details are shown in Figure \ref{code:vit-1k}.
\section{Details of generating \textsc{GPTs-1K} dataset}
\label{appendix_gpts1k}
We also change the values in layers $L$, heads $H$ and hidden size $D$ of GPT-2, and the details are shown in Figure \ref{code:gpt-1k}.

\end{document}